\documentclass[conference]{IEEEtran}
\usepackage{times}

\usepackage[numbers]{natbib}
\usepackage{multicol}
\usepackage[bookmarks=true]{hyperref}

\usepackage{graphicx}
\usepackage{amsmath}
\usepackage{tikz}
\usetikzlibrary{bayesnet}
\usetikzlibrary{arrows}
\usepackage{bbm, dsfont}
\usepackage{multicol}
\usepackage{lipsum}

\usepackage[font=small,labelfont=bf]{caption}

\usepackage[utf8]{inputenc} 
\usepackage[T1]{fontenc}    
\usepackage{hyperref}       
\usepackage{url}            
\usepackage{booktabs}       
\usepackage{amsfonts}       
\usepackage{nicefrac}       
\usepackage{microtype}      
\usepackage{xcolor}         

\usepackage{algpseudocode}
\usepackage{algorithm}

\usepackage{subcaption}

\begin{document}

\title{Semi-Supervised Neural Processes for Articulated Object Interactions}



\author{
\centering
Emily Liu$^1$
 \and Michael Noseworthy$^1$\\
$^1$Massachusetts Institute of Technology
\and Nicholas Roy$^1$
}

\maketitle

\begin{abstract}


The scarcity of labeled action data poses a considerable challenge for developing machine learning algorithms for robotic object manipulation. It is expensive and often infeasible for a robot to interact with many objects. Conversely, visual data of objects, without interaction, is abundantly available and can be leveraged for pretraining and feature extraction. However, current methods that rely on image data for pretraining do not easily adapt to task-specific predictions, since the learned features are not guaranteed to be relevant. This paper introduces the Semi-Supervised Neural Process (SSNP): an adaptive reward-prediction model designed for scenarios in which only a small subset of objects have labeled interaction data. 
In addition to predicting reward labels, the latent-space of the SSNP is jointly trained with an autoencoding objective using passive data from a much larger set of objects.
Jointly training with both types of data allows the model to focus more effectively on generalizable features and minimizes the need for extensive retraining, thereby reducing computational demands.
The efficacy of SSNP is demonstrated through a door-opening task, leading to better performance than other semi-supervised methods, and only using a fraction of the data compared to other adaptive models.

\end{abstract}

\IEEEpeerreviewmaketitle

\section{Introduction}

In this work, we are interested in learning manipulation models that are \emph{adaptive}: as a robot interacts and observes outcomes for a novel object, its performance should improve.
Recently, a class of models called \emph{Neural Processes} have been proposed that achieve this goal.
However, they typically require labeled interactions for a large set of objects that is difficult to obtain \citep{neuralprocesses} \citep{amortizedinferencegrasp}.
Interacting with many different objects is costly in terms of both time and money.
For a fixed manipulator, the robot's workspace can only fit a limited number of objects.
Even for mobile manipulators, moving to new objects can take an infeasible amount of time.

However, other types of data, such as images or video, are much more readily available for a diverse set of objects.
Robots are commonly equipped with cameras which allow them to passively observe the interactions of other agents: for example, it can watch a human opening a door.
There are also a wide range of video and image data on the web of either humans or other robots interacting with the world \cite{Damen2018EPICKITCHENS}.
We are interested in models for manipulation that use labeled data generated from fewer objects by leveraging widely available passive data.


Concretely, we assume the data are organized by objects, where all objects have multiple associated images, but only a fraction of the objects have action data labeled with outcomes (rewards).
Manipulation models trained only on the objects that include labels (i.e., supervised learning) are unlikely to generalize to new objects due to the poor coverage of the dataset.
In such scenarios, previous works have considered semi-supervised setups to leverage additional unlabeled data.

A common approach is a pretraining-finetuning approach, where the model is first pretrained on unlabeled data (i.e., images) and then finetuned on labeled data (i.e., action-reward pairs) \citep{kumar2023pretraining} \citep{dipalo2024dinobot} \citep{aoyama2023fewshot}. One issue with pretraining is domain shift, as the learned features may not be useful for the downstream manipulation task. Other techniques \citep{iwata2023metalearning} \citep{thomas2023plex} have been proposed to train both labeled and unlabeled data simultaneously without requiring pretraining, but do not account for the structured data common in manipulation tasks (i.e., datasets are hierarchical with many actions for each object). 


To develop adaptive models using labeled data from fewer objects, this work proposes the Semi-Supervised Neural Process (SSNP) model for object manipulation. More concretely, the model predicts the outcome of a manipulation action, given a set of object images and outcomes from any previous interactions. To leverage the larger set of objects with only unlabelled videos or images, the latent-space of the SSNP is jointly trained with an autoencoding objective for images and actions. A key ingredient to this approach is that instead of having separate pretraining and finetuning steps, we train using labeled and unlabeled data simultaneously. We argue that training on actions simultaneously helps the image autoencoder focus on features important to the manipulation task. Because the model does not require additional training epochs for finetuning, the computational cost of training is also greatly reduced. We demonstrate the method using a door opening task, showing that compared to baselines, the model is better able to leverage relevant features from images to robustly predict action outcomes and adapt to previous data in a few-shot manner.




\begin{figure}
    \centering
\includegraphics[width=0.38\textwidth]{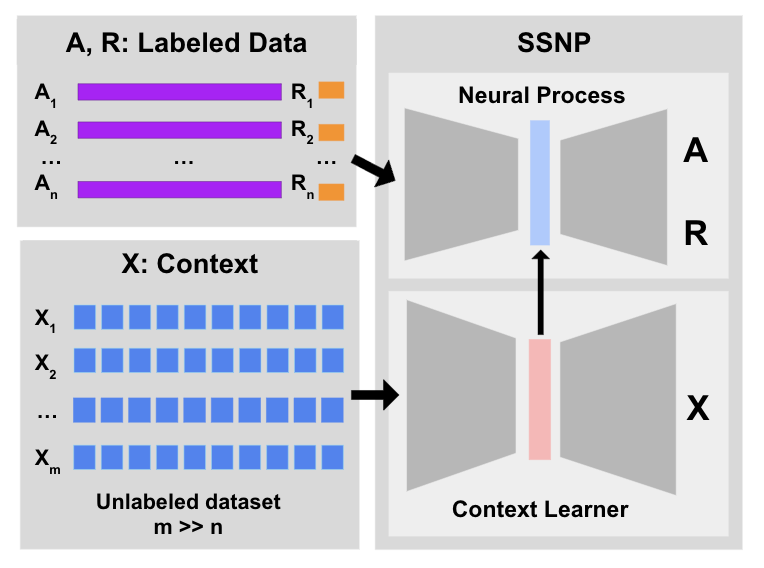}
    \caption{The Semi-Supervised Neural Process architecture consists of two autoencoder components. The context learner learns embeddings for unlabeled data (such as images) and the neural process learns an embedding of labeled action-reward pairs.}
    \label{fig:modeldiagram}
\end{figure}

\begin{figure*}[t]
    \centering
\includegraphics[width=0.9\textwidth]{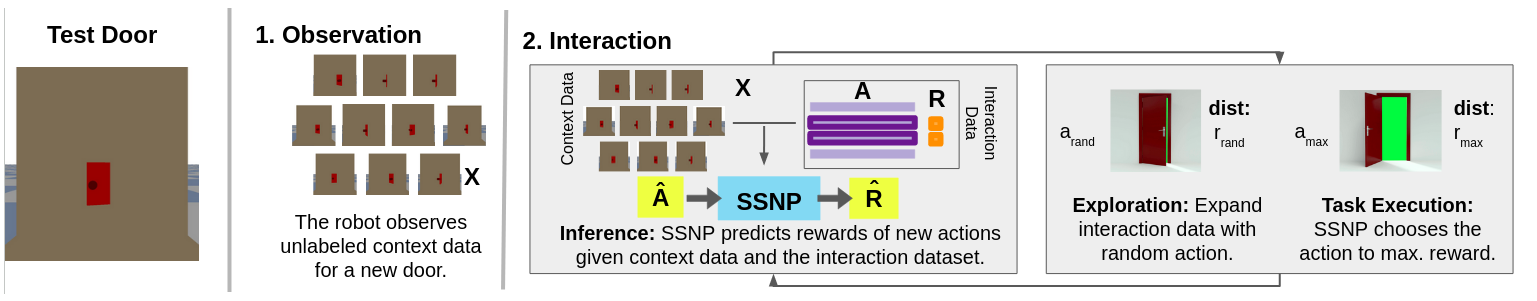}
    \caption{Overview of the Semi-supervised Neural Process in deployment. The robot is presented with a new object with unknown kinematic properties. (1)The robot can passively observe \emph{context data} without interacting with it. (2) When the robot starts interacting with the object, it can iteratively build an \emph{interaction dataset} (exploration). The SSNP model can also be used to predict optimal actions based on the limited interaction data (task execution).}
    \label{fig:overview}
\end{figure*}

\section{Problem Formulation and Background}

\begin{figure*}[t]
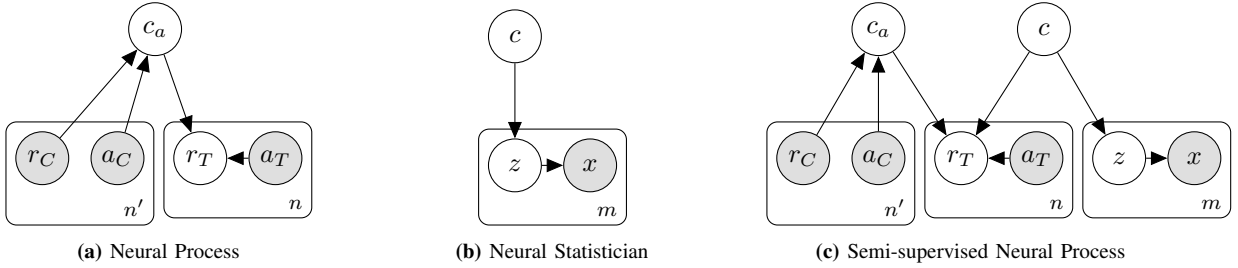

    \centering
    
    \begin{subfigure}[t]{0.25\textwidth}
        \centering
        \tikz{
 \node[obs, xshift=-3cm] (ac) {$a_C$};%
 \node[obs, xshift=-4cm] (rc) {$r_C$};%
 \node[obs, xshift=-0.9cm] (at) {$a_T$};%
 \node[latent, xshift=-1.9cm] (rt) {$r_T$};%
 \node[latent, above=of ac, xshift=0.5cm] (ca) {$c_a$};%
 \edge {ac, rc} {ca} 
 \edge {at, ca} {rt}
 
\plate {} {(ac) (rc)} {$n'$}; 
\plate {} {(at) (rt)} {$n$}; 
}
        \caption{Neural Process}
    \end{subfigure}%
    ~ 
    \begin{subfigure}[t]{0.3\textwidth}
        \centering
        \tikz{

 \node[latent, xshift=5cm] (z) {$z$};%
 \node[obs, xshift=6cm] (x) {$x$};%
 \node[latent, above=of z, xshift=0cm] (c) {$c$};
 \edge {c} {z} 
 \edge {z} {x}
 
\plate {} {(z) (x)} {$m$}; 
 }
        \caption{Neural Statistician}
    \end{subfigure}
    \begin{subfigure}[t]{0.3\textwidth}
        \centering
\tikz{
 \node[latent, xshift=1.2cm] (z) {$z$};%
 \node[obs, xshift=2.2cm] (x) {$x$};%
 \node[latent, above=of z, xshift=-1cm] (c) {$c$};
 \edge {c} {z} 
 \edge {z} {x}
 
\plate {} {(z) (x)} {$m$}; 
 \node[obs, xshift=-2cm] (ac) {$a_C$};%
 \node[obs, xshift=-3cm] (rc) {$r_C$};%
 \node[obs, xshift=0.1cm] (at) {$a_T$};%
 \node[latent, xshift=-0.9cm] (rt) {$r_T$};%
 \node[latent, above=of ac, xshift=0cm] (ca) {$c_a$};%
 \edge {ac, rc} {ca} 
 \edge {at, ca, c} {rt}
 
\plate {} {(ac) (rc)} {$n'$}; 
\plate {} {(at) (rt)} {$n$}; 
 }
        \caption{Semi-supervised Neural Process}
    \end{subfigure}
    
    \caption{Bayesian graphical models for NP, NS, and SSNP. \textbf{(a)} In the Neural Process, the observed variables are the pre-existing action-reward pairs ($a_C$, $r_C$) and the actions with unseen rewards ($a_T$). We learn the associated reward $r_T$ with $a_T$ by means of a latent variable $c_a$. \textbf{(b)} In the Neural Statistician, we observe hierarchical data $x$ (e.g., images), from which we learn shared contextual latent variables $c$ corresponding to the object, and unique instance-level latent variables $z$ for each individual sample. \textbf{(c)} The SSNP integrates the NS and NP probabilistic generative models, where the unobserved $r_T$ is conditioned upon both action-level latent variables $c_a$ and object-level latent variables $c$.}
    \label{fig:graphicalmodels}
\end{figure*}

Our goal is to create an adaptive model for predicting action outcomes.
We focus on a scenario where the dataset is comprised of objects annotated with multiple types of data.
Each object has unlabeled data that is easy to obtain (e.g., image data), and a set of labeled data which is more costly to obtain (e.g., action-reward pairs).
We assume that only a fraction of the objects includes the labeled data.

\subsection{Notation}
SSNP solves a model-based bandit task and handles hierarchical multimodal datasets where each object $O_i$ in the dataset $\mathcal{D}$ has two associated sets of data, $X_i = \{x_i^1 ... x_i^m\}$ and  $A_i = \{a_i^1 ... a_i^n\}$ with $m$ and $n$ individual datapoints respectively. 
Each element of $A_i$ is an action (we assume all actions are performed from some nominal state), and each element of $X_i$ is additional context data relevant to the object. 
$A_i$ is also associated with a set of labels, $R_i = \{r_i^1, ... r_i^{n'}\}$, that may be incomplete ($0 \leq n' \leq n$) and corresponds to the reward associated with applying the respective action. 

In this work, we consider a door opening domain.
Each object is a door with different, and unknown, kinematic parameters (e.g., location of rotation axis, distance of handle to axis).
$X$ is a set of images of the door opened at different angles (which could be passively observed).
$A$ is a set of different parameterized trajectories that a manipulator could use to open the door from closed.
$R$ would be the real-valued distance travelled by the handle after the action is applied (note that maximum reward is only achieved if the chosen trajectory aligns with the object's unknown kinematic properties otherwise the gripper may jam or slip off the handle).

\subsection{Learning Phases}
\textbf{Training}: First, we assume there is a distinct offline training phase where a model can be learned from a fixed dataset covering many objects.
Motivated by the problem setup, we additionally assume that either all the actions for an object are labeled, or none of them are.
We are interested in scenarios where most of the objects \emph{only} have unlabeled data (equivalently, no interaction data).
In the doors domain, unlabeled images are easy to obtain (e.g., from online datasets or a robot navigating a building), but it is costly to interact with a large variety of doors.

\textbf{Inference}:
After training, the robot many encounter a new object and should be able to efficiently interact with it.
Initially, it will observe only unlabeled images of the new object.
As it interacts with the object and observes outcomes, these results can also be given to the model to improve its predictions.
We are interested in models that allow this \emph{adaptive} capability.


\subsection{Neural Processes}
\label{sec:background-np}

A \emph{Neural Process} (NP) is a class of models that allows this adaptive behaviour \citep{neuralprocesses}.
NPs are inspired by VAEs and Gaussian Processes, allowing efficient inference over functions (in our case, inference over reward-prediction functions).
In the inference phase, given contextual data about object $i$ (e.g., images $X_i$) and a set of attempted actions and their outcomes ($A_i, R_i$), a NP can predict reward values for new actions that are consistent with the observed data so far.
Importantly, the set of labeled actions can have a variable size, leading to predictions that become increasingly accurate as more interactions occur.

The model itself consists of an \emph{encoder}, which aggregates previous interaction data and contextual data, and a \emph{decoder}, which predicts reward values given new actions and the output of the encoder.
The model is trained in an auto-encoding fashion and uses a second forward pass through the encoder, with subset of the labeled actions, to match the latent variables learned on the full action set and on the partial action set.

A limitation of current approaches that use NPs for robotic manipulation is they assume a fully labeled dataset in the training phase: that is, all objects contain labeled actions. 
In this paper, we are interested in whether we can use passive data such as images or videos to reduce this burden.


\section{Semi-Supervised Neural Processes}
Our key contribution is to extend the NP framework to scenarios where many objects do not have reward data.
We propose \emph{Semi-Supervised Neural Processes} (SSNPs) which are composed of two autoencoder-like modules: the original NP (see Section \ref{sec:background-np}) and a \emph{context-learner}. 
The \emph{context-learner} is a set-based image auto-encoder trained to reconstruct a set of images all belonging to the same object.
Features from the encoder of the \emph{context-learner} are given to the NP's decoder to make predictions about action rewards.
Because the \emph{context-learner} can be learned from a larger dataset with more objects (it does not require labeled actions), we expect it to learn generalizable features that also help with reward-prediction.

\subsection{Model Overview}

The SSNP consists of three modules: the \emph{context-learner}, the \emph{actions encoder}, and the \emph{actions decoder}.
Figure \ref{fig:modeldiagram} shows the overall architecture of the model. 


\textbf{Context Learner}
The context learner encodes all the unlabeled context data  (e.g., images) associated with the object $O$. 
In the case of the door example, the context learner takes images, $X$, of a door opened at varying configurations and generates a latent variable $c$ that captures features shared across elements of $X$.
We draw inspiration from the Neural Statistician model \citep{edwards2017neural} for this component of our model.
Instead of reconstructing images independently, images belonging to the same door are reconstructed together using a single set of latent variables.
This allows the \emph{context-learner} to extract features that contain object-level information that may not be observable from a single image, yet are important for interactive tasks such as door opening.


\textbf{Actions Encoder}
The \emph{actions encoder} is the same as in the NP model but will be trained with less labeled data.

\textbf{Actions Decoder}
The \emph{actions decoder} takes in the output of the \emph{context-learner's} encoder, as well the the output of the \emph{actions encoder}.
Given a new action, it aims to predict the reward for this action.

\subsection{SSNP Training Procedure}

The semi-supervised training process is similar to the fully supervised NP, with the exception that most batches do not have labeled action data.
When only unlabeled image data is included in a batch, the context-learner is updated using the Neural Statistician loss function:
\begin{align*}
\mathcal{L}_X =& \, \mathbbm{E}_{q(c | X)} [\mathbbm{E}_{q(z | c, x)} \left[\log p(x | z)\right] \\ & - D_{KL}(q(z | c, x) || p(z | c))] - D_{KL}(q(c | X) || p(c)).
\end{align*}
Along with a context-level latent variable, $c$, shared across images, we also include an image-level latent variable, $z$, to capture nuisance features like camera placement. These features are ignored by the action decoder. For more details, see the NS paper \cite{edwards2017neural}.

When batches also have interaction data (i.e., $(A, R)$),  all components of our model are updated using $\mathcal{L}_{tot}$, which updates the \emph{context-learner} through $\mathcal{L}_X$, and the \emph{action auto-encoder} through $\mathcal{L}_a$:
\begin{align*}
    \mathcal{L}_{tot} =& \,   \mathcal{L}_a +   \mathcal{L}_X \\
    \mathcal{L}_a =& \, MSE(\hat{R}, R) +D_{KL} (q_{action}(c_a | A, R) || N(0, 1)) \\ &+ D_{KL} (q_{action}(c_a' | A', R') || q_{action}(c_a | A, R)) 
\end{align*}
Because SSNP makes inferences even with partial labels, we include a second forward pass using a subset of the labeled actions ($A', R'$) to match the latent variables inferred from the full available action set ($A, R$). We reflect this in the variational lower bound, $\mathcal{L}_a$, by including the KL divergence between the partial posterior $q_{action}(c_a' | A', R')$ and the full posterior $q_{action}(c_a | A, R)$. The size of the subset is randomly sampled.

\subsection{SSNP Inference}

When presented with a novel object, the SSNP is used equivalently to an NP.
The inclusion of unlabeled object data leads to better performance during this phase.



\begin{figure*}[t]
    \centering
        \includegraphics[width=0.25\textwidth]{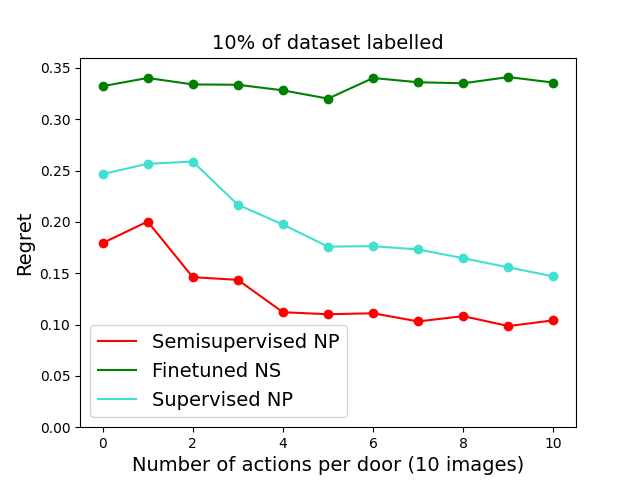}
    \includegraphics[width=0.25\textwidth]{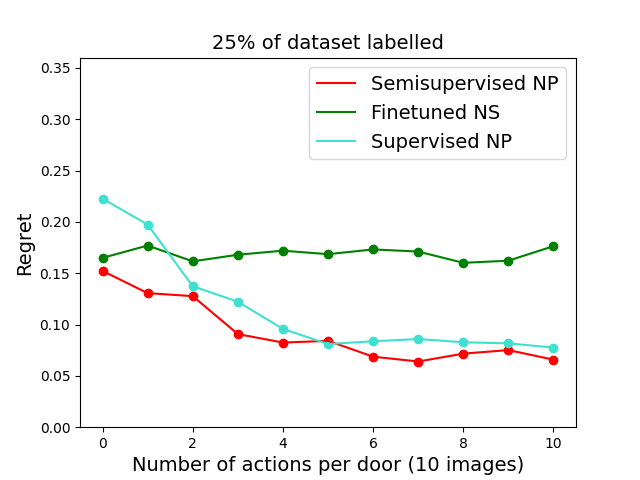}
    \includegraphics[width=0.25\textwidth]{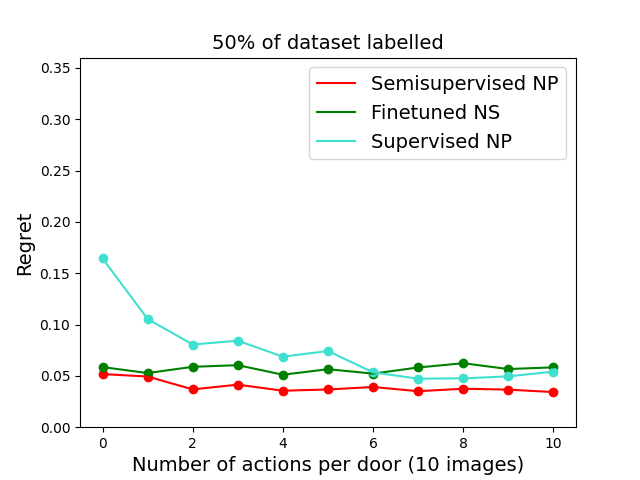}
    \caption{
    Semi-Supervised Neural Process vs Neural Statistician and Neural Process baselines, for datasets with 10\%, 25\%, and 50\% labeled action data. The NS baseline (Finetuned NS) improves with the fraction of labeled data but is unable to adapt to any specific object. The NP baseline's performance degrades with smaller labeled datasets. On the other hand, SSNP exhibits adaptive behaviour and achieves good performance, even with few labeled objects. Regret is evaluated over 100 random actions. The number of labeled actions is capped at 10.}
    \label{fig:baselinecomps}
\end{figure*}

\section{Experiments}
In this section, we present an in-depth analysis on the efficacy of the model reward prediction and adaptability across different levels of dataset supervision. We find that SSNP achieves lower error and is more adaptable compared to baselines.


\subsection{Dataset}\label{dataset}
We use a synthetic dataset of doors generated by PyBullet. Our data consists of 8000 doors in the training set, and 200 in the test set. Each door is represented by $m$ images ($m$ =10) and $n$ actions ($n$ = 10). For a single door, the image data consists of snapshots taken at distances between 20 and 40 cm. The door orientation ranges from closed to open 180 degrees. Each action is parameterized by the axis and location of the rotation center, the distance between the point of contact and the axis of rotation, and the goal configuration of the object \citep{cpp}. Each action is additionally paired with a reward, the displacement of the handle after revolution around the center of rotation if the action policy is executed. We simulate a semi-supervised setup in which only a given percentage $k$ of doors in the training dataset are labeled ($k = [10, 25, 50]$).

\subsection{Baselines}
\textbf{Supervised Neural Process} 
Our first baseline is a Neural Process that only utilizes labeled data. This is equivalent to the SSNP ignoring the unlabeled data.

\textbf{Neural Statistician}
The Neural Statistician (NS) \citep{edwards2017neural} is a latent variable model designed to perform inference on sets of data points, allowing the model to learn latent variables that encapsulate features shared by set elements. The NS architecture lends itself well to the doors domain: using a pretrained NS allows for effective finetuning on a small labeled dataset. However, the process of finetuning is costly and requires additional compute and training time on top of the time it already took to pretrain the neural statistician. 
In our problem set-up, we pretrain the neural statistician on the images alone, and then use the pretrained embeddings to learn action-reward pairs using a fully connected network. 
This fine-tuning is done with the labeled data.

\subsection{Evaluations}
When the robot is presented with a new object, we want to measure how well it performs throughout the inference phase as it collects more information about the object.
As such, we present metrics as we increase the number of actions per door.
To assess the quality of reward predictions, we use regret evaluations on novel doors.
%
For each test door, we report regret by using the rewards predicted by the model to select the best action-reward pair ($\hat{a}$, $\hat{r}$) out of $100$ randomly generated actions.
The reward from the selected action is compared to the reward $r^*$ associated with the optimal action $a^*$ for the object. The regret is given by $\frac{r^* - \hat{r}}{r^*}.$

We also evaluate the RMSE of the reward predictions which show similar trends to the regret metric. More details can be found in the appendix.



\subsection{Results and Discussion}
Our results show that SSNP performs robustly when trained with labeled data from fewer objects. 
Figure \ref{fig:baselinecomps} compares SSNP to NP and finetuned NS baselines on datasets with 10\%, 25\%, and 50\% supervision, in order to simulate a real-world scenario in which the majority of data are unlabeled. 

When compared to the fully supervised NP baseline, the SSNP model achieves better adaptive performance when using the same number of doors with labeled data.
Without any labeled interactions for the test door (i.e., $x=0$ in Figure \ref{fig:baselinecomps}), SSNP always outperforms the NP method.
In this case, the models are making predictions only from images.
We also see that the SSNP quickly achieves lower than the NP baseline when the number of labeled actions increases. 
This is especially apparent in the regret plots, as the SSNP reaches a ``stopping'' point at around 4 actions for the 10\% and 25\% labeled datasets, but the NP continues to converge as more actions are added. Thus, the image autoencoder of the SSNP also allows for more efficient learning given a new door.
Overall the SSNP can better generalize to test doors than the NP because it can leverage data from a wider set of unlabeled objects.


When compared to the finetuned NS baseline, we notice comparable performance with the SSNP model when no adaptation interactions have occured ($x=0$).
However, the NS baseline has no ability to adapt: it does not incorporate the outcomes of actions with the test door to further improve its predictions. 
The SSNP, on the other hand, is able to improve because of its latent space matching component previously discussed.
In addition, the SSNP model only uses a single training phase, compared to the NS baseline which using both a pre-training and fine-tuning phase. 
This results in a simpler and faster training procedure.

\section{Conclusion}
In this paper, we introduce the Semi-Supervised Neural Process (SSNP) model. 
The integration of a shared context-level autoencoder and a neural process framework within the SSNP model allows for an effective semi-supervised learning strategy that generalizes efficiently to new objects and significantly reduces the need for extensive labeled datasets. This approach not only improves the focus on manipulation-relevant features but also diminishes computational overhead by minimizing the necessity for subsequent retraining phases.
We demonstrate the practicality of SSNP with a door-opening task, where it outperforms supervised and pretrained baselines even when only 10\% of the objects in the dataset have labels.



\nocite{aoyama2023fewshot}
\nocite{s3k}
\nocite{iwata2023metalearning}
\nocite{kingma2022autoencoding}
\nocite{CNP}
\nocite{amortizedinferencegrasp}
\nocite{cvae}
\bibliographystyle{unsrtnat}
\bibliography{references}

\newpage
\section{Appendix}

\subsection{Model Details}
Context learner:
\begin{enumerate}
    \item \textbf{Set Encoding ($X \rightarrow h$)}: All elements of $X$ are passed into a shared encoder. In the case of image data, the set encoder is a shared convolutional encoder. The output $h$ is an embedding that incorporates each separate data point of $X$.
    \item \textbf{Statistic Network ($q(c | h; \phi)$)}: The statistic network learns a latent variable $c$ from $h$. We $h$ through a fully connected network to learn a normal distribution with a unit normal prior, from which $c$ is sampled.
    \item \textbf{Inference Network ($q(z | c, h; \phi)$)}: The purpose of the inference network is to ensure that $c$ only captures information relevant to the object across \textit{all} images, by capturing the instance-level parameters in a second latent variable $z$ conditioned on $c$. The conditional distribution of $z$ is learned similarly to that of $c$.
    \item \textbf{Latent Decoder ($p(z | c; \theta)$)}: The latent decoder learns an appropriate prior for each $z$ variable, independent of the input $h$. During the computation of the variational lower bound, the distribution learned by the inference network is made to match that of the latent decoder.
    \item \textbf{Observation Decoder ($p(X | c, z; \theta)$)}: The observational decoder  reconstructs the original set of images $X$.
\end{enumerate}
Action model:
\begin{enumerate}
    \item \textbf{Set Encoder ($(A', R') \rightarrow h_a$)}: Similar to the context learner, the first step of the neural process is to aggregate all relevant labeled data into a single vector.
    \item \textbf{Aggregator ($q(c_a | h_a; \phi)$)}: The neural process learns a shared latent space for all labeled data pairs. $c_a$ is sampled from a normal distribution with a prior of $N(0, 1)$, and concatenated with $c$ to include contextual information in a shared latent variable.
    \item \textbf{Conditional Decoder ($p(R^* | c_a, c, A; \theta)$)}: The conditional decoder predicts the unobserved label distribution conditioned on the latent space of encoded data-label pairs along with a set of unlabeled target data.
\end{enumerate}
\subsection{Training/inference algorithms}
See: Algorithm \ref{alg:training}, Algorithm \ref{alg:trainingsemi}, Algorithm \ref{alg:inference}. Both image and action objective functions were trained using a KL annealing scheme.





\begin{algorithm}
\caption{Training (Semi-supervised)}\label{alg:trainingsemi}
\begin{algorithmic}[1]
\Require Image data $\mathcal{X} \in \mathbbm{R}^{D \times m}$, Action data $\mathcal{A} \in \mathbbm{R}^{D \times n}$, Reward data $\mathcal{R} \in \mathbbm{R}^{D \times n}$ in a dataset of size $D$

\For{$t = 1 ... T$}
\State $X, a, r \gets \mathcal{X}_i, \mathcal{A}_i, \mathcal{R}_i$

\State $q_{img}(c | X), q_{img}(z | c, X) \gets enc_{img} (X)$
\State $p_{img}(\hat{X} | z, c) \gets dec_{img} (c_X, z_X)$

\State $\mathcal{L}_X \gets \mathbbm{E}_{q(c | X)} [\mathbbm{E}_{q(z | c, x)} \left[\log p(x | z)\right] $ \State $ \quad\quad- D_{KL}(q(z | c, x) || p(z | c))] - D_{KL}(q(c | X) || p(c))$

\If{$r \neq \emptyset$}
\State $q_{action}(c | a) \gets enc_{action} (a, r)$
\State $\hat{r} \gets dec_{action} (c, c_a, a)$
\State let $m' \gets rand(1, m)$
\State $q_{action}(c_a' | a) \gets enc_{action} (a[:m'], r[:m'])$

\State $\mathcal{L}_a \gets MSE(\hat{r}, r) +D_{KL} (q_{action}(c_a | a) || N(0, 1))$
\State$\quad\quad\quad + D_{KL} (q_{action}(c_a' | a) || q_{action}(c | a))$
\EndIf
\If {$r = \emptyset$}
\State $\mathcal{L}_a \gets 0$
\EndIf
\State $\mathcal{L} \gets   \mathcal{L}_a +   \mathcal{L}_X  $

\EndFor
\end{algorithmic}
\end{algorithm}

\begin{algorithm}
\caption{Inference}\label{alg:inference}
\begin{algorithmic}[1]
\Require Image data $\mathcal{X} \in \mathbbm{R}^{D \times m}$, Action data $\mathcal{A} \in \mathbbm{R}^{D \times n}$, Reward data $\mathcal{R} \in \mathbbm{R}^{D \times n'}$ in a dataset of size $D$ where $n' \leq n$

\State $X, a, r \gets \mathcal{X}_i, \mathcal{A}_i, \mathcal{R}_i$
\State Let $a' = a[n':]$ be the set of labeled actions
\State $q_{img}(c | X), q_{img}(z | c, X) \gets enc_{img} (X)$
\State $p_{img}(\hat{X} | z, c) \gets dec_{img} (c_X, z_X)$
\State $q_{action}(c | a) \gets enc_{action} (a', r)$
\State $\hat{r} \gets dec_{action} (c, c_a, a)$
\State Return $\hat{r}$

\end{algorithmic}
\end{algorithm}
\subsection{Evaluation on Number of images per door}

\begin{figure}[H]
    \centering
    \includegraphics[width=0.24\textwidth]{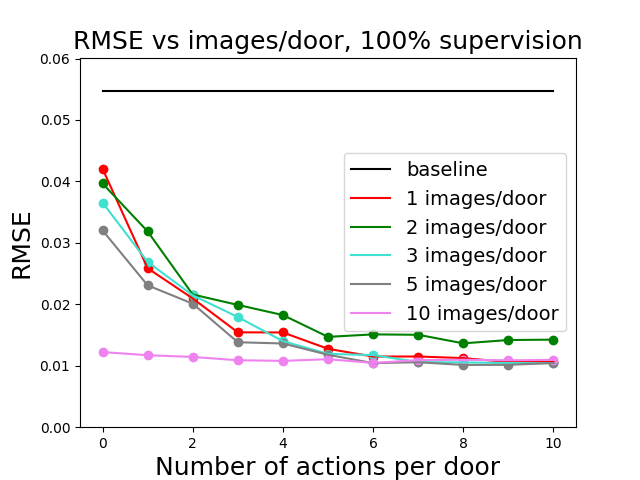}
    \includegraphics[width=0.24\textwidth]{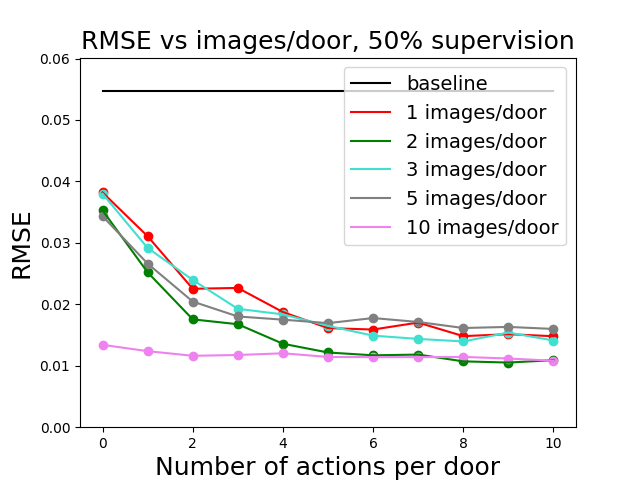}
    \includegraphics[width=0.24\textwidth]{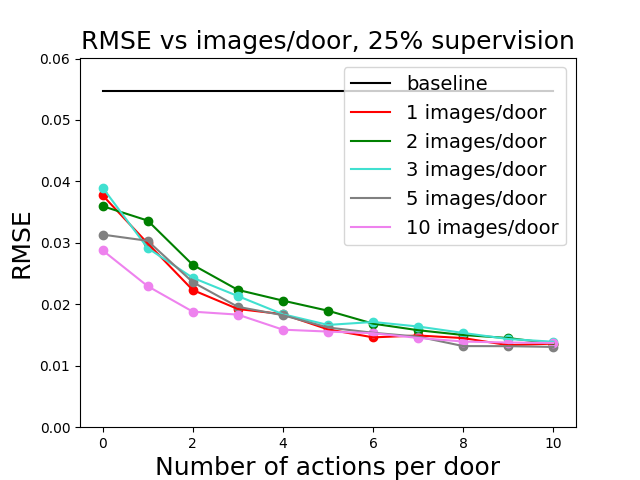}
    \includegraphics[width=0.24\textwidth]{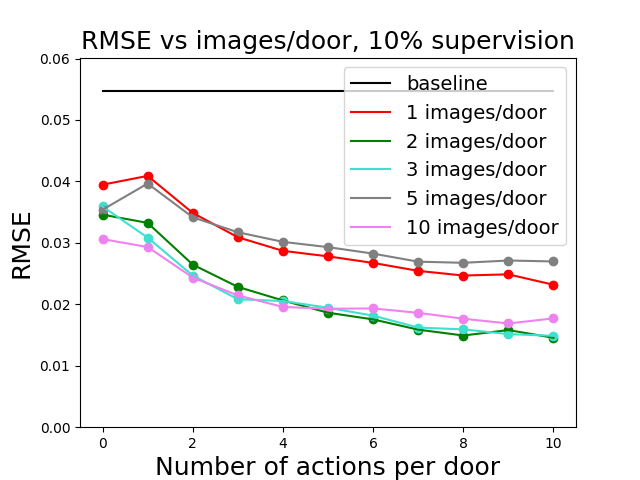}
    \caption{Root mean squared error on reward prediction on SSNP model for different numbers of actions, comparing across different levels of supervision and number of images in the context autoencoder. Having more images in the autoencoder benefits predictions even with low data supervision. The baseline (black line) is the standard deviation of the rewards in the test dataset.}
    \label{fig:mask_comparisons}
\end{figure}
Figure \ref{fig:mask_comparisons} shows the effect of different levels of supervision on reward prediction error, measured in root mean squared error (RMSE). We observe that for 1-5 images per object, making predictions with a small number of labeled actions but reaches lower values as more labels are observed. When we increase the number of images per object to 10, we notice that the model is able to make accurate inferences with low RMSE even when no previous actions are seen. 
The effect is most noticeable when the dataset is fully labeled, but even when the training set is partially labeled at 50\%, 25\%, and 10\% of doors, we observe similar trends. The results observed here are in line with our expectations, since the model is able to learn predictions with low error even with extremely low amounts of labeled data. Even though the error increases as the percentage of labeled training doors decreases, our model still outperforms baselines at low levels of data supervision.

\subsection{RMSE comparisons with baselines}
\begin{figure}
    \centering
        \includegraphics[width=0.3\textwidth]{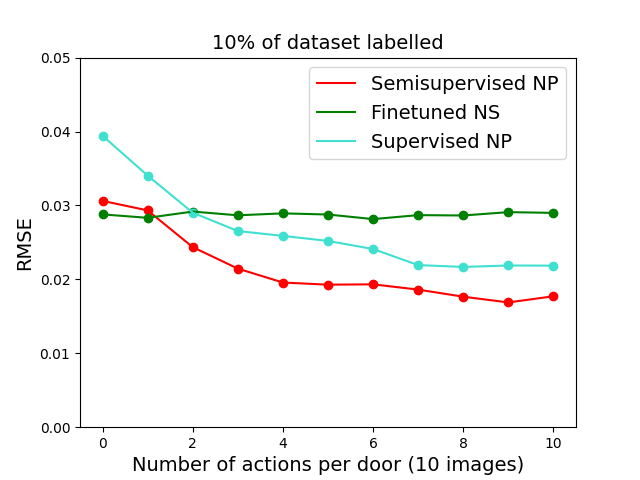}
    \includegraphics[width=0.3\textwidth]{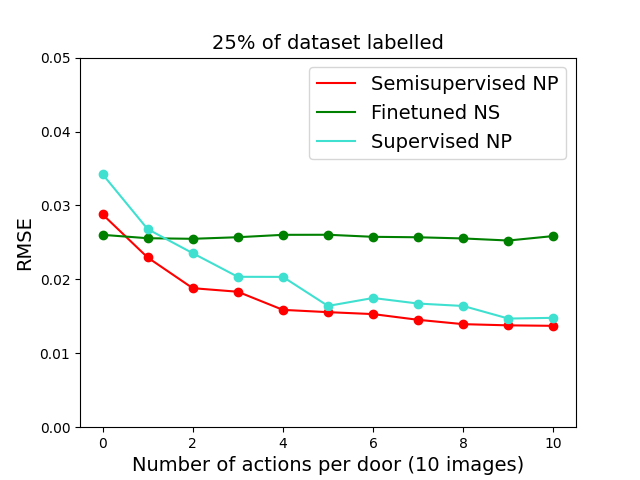}
    \includegraphics[width=0.3\textwidth]{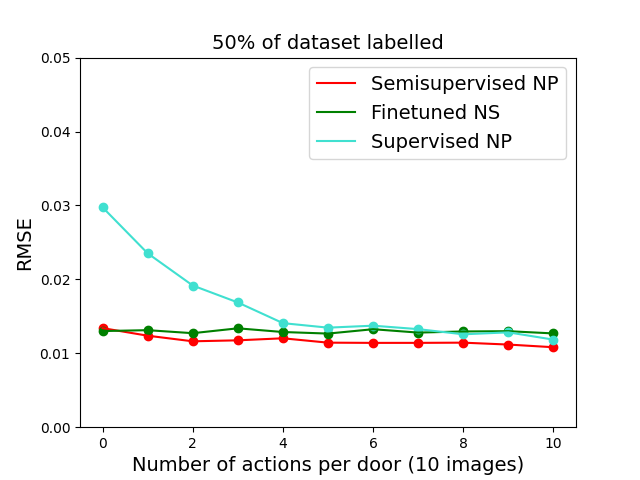}

    \caption{Root mean squared error}
    \label{fig:enter-label}
\end{figure}

Figure \ref{fig:enter-label} shows the same comparisons as Figure \ref{fig:baselinecomps}, done with reward RMSE in place of regret. RMSE is averaged across $10$ actions for a given object, and the average RMSE is reported across $200$ test doors. 
\end{document}